\documentclass[10pt,twocolumn,letterpaper]{article}

\usepackage{3dv}
\usepackage{times}
\usepackage{epsfig}
\usepackage{graphicx}
\usepackage{amsmath}
\usepackage{amssymb}
\usepackage{enumitem}
\usepackage{mathrsfs}
\usepackage{booktabs}
\usepackage{multirow}
\usepackage{fancyhdr}

\fancypagestyle{plain}{
	\fancyhf{}
	\fancyhead[C]{Fifth International Conference on 3D Vision 3DV, 2017}    
	\fancyfoot[C]{\thepage}%

}
\usepackage[pagebackref=true,breaklinks=true,letterpaper=true,colorlinks,bookmarks=false]{hyperref}

\newcommand{\argmin}{\arg\!\min}
\threedvfinalcopy 

\def\threedvPaperID{8} 

\ifthreedvfinal\fi
\begin{document}

\title{4D Temporally Coherent Light-field Video}

\author{Armin Mustafa  \hspace{.07\linewidth} Marco Volino \hspace{.07\linewidth} Jean-Yves Guillemaut \hspace{.07\linewidth} Adrian Hilton \\
	CVSSP, University of Surrey, United Kingdom\\
	{\tt\small \{a.mustafa, m.volino, j.guillemaut, a.hilton\} @surrey.ac.uk}
}

\maketitle

\begin{abstract}
Light-field video has recently been used in virtual and augmented reality applications to increase realism and immersion. However, existing light-field methods are generally limited to static scenes due to the requirement to acquire a dense scene representation. The large amount of data and the absence of methods to infer temporal coherence pose major challenges in storage, compression and editing compared to conventional video. In this paper, we propose the first method to extract a spatio-temporally coherent light-field video representation. A novel method to obtain Epipolar Plane Images (EPIs) from a spare light-field camera array is proposed. EPIs are used to constrain scene flow estimation to obtain 4D temporally coherent representations of dynamic light-fields. Temporal coherence is achieved on a variety of light-field datasets. Evaluation of the proposed light-field scene flow against existing multi-view dense correspondence approaches demonstrates a significant improvement in accuracy of temporal coherence.
\end{abstract}
%
%
\vspace{-0.75cm}
\section{Introduction}
Light-field cameras capture multiple densely spaced viewpoints using either lenticular arrays for a single sensor \cite{Levoy2006}, a camera densely scanned over a static scene to capture a large number of views \cite{Kim2013} or multiple densely spaced cameras \cite{Wilburn2005}. This allows photo-realistic rendering of novel viewpoints, depth-of-field effects and high-accuracy dense reconstruction \cite{Kim2013, Vaish06}. Reconstruction from densely sampled light-fields commonly employs an Epipolar Plane Image (EPI) representation to estimate spatial correspondence between images \cite{Yucer2016}.

\begin{figure}
	\centering
	\includegraphics[width=0.99\linewidth]{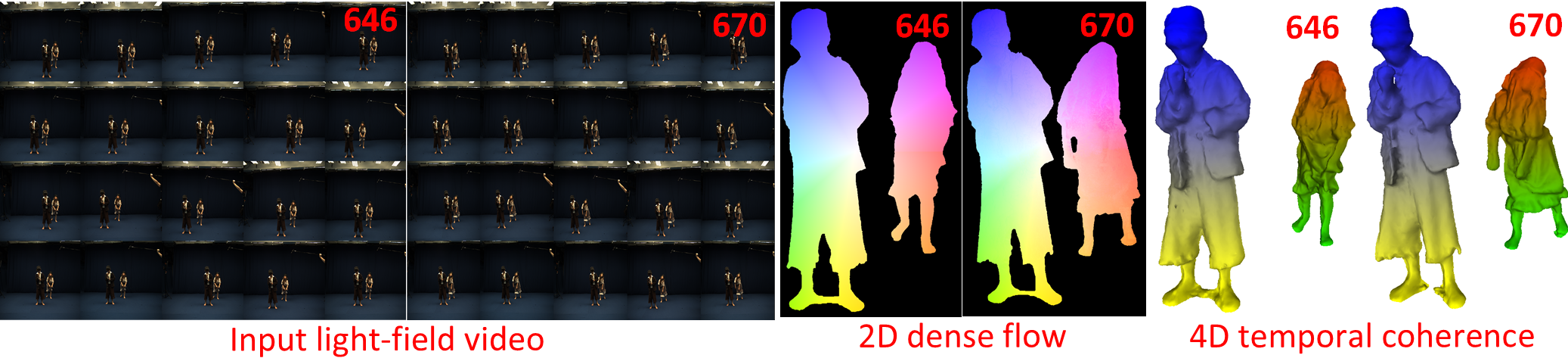}
	\caption{Proposed 4D temporally coherent dynamic light-fields}
	\label{fig:motivate}
\end{figure}

\noindent
Light-fields have primarily been captured for static scenes due to the requirement for a large number of viewpoints resulting in a high-volume of data.  
Recently  motivated by applications in immersive content production for Virtual and Augmented Reality (VR/AR), arrays of video cameras could be employed to acquire light-fields of dynamic scenes.
However, conventional video editing techniques are not suitable to edit such data as they fail to exploit the spatial and temporal redundancy encoded in light-fields due to multiple discrete views.
Hence there is a need for efficient light-field video representations exploiting the spatio-temporal redundancy to enable compression and facilitate editing for live action VR.
In this paper we propose the first method for spatio-temporally coherent dynamic light-field video, illustrated in Figure \ref{fig:motivate}. This enables the use of light-field video for immersive VR content production. An EPI representation is used to estimate dense spatio-temporal correspondence from light-field video for temporal alignment.
\begin{figure*}[t]
	\centering
	\includegraphics[width=0.97\linewidth]{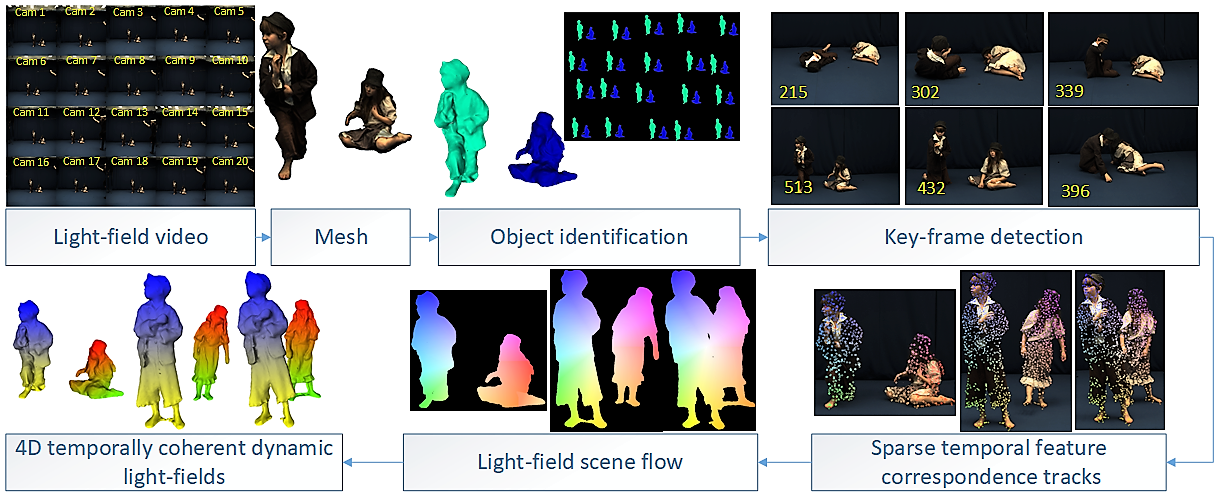}
	\vspace{-0.2cm}
	\caption{4D Temporally Coherent Light-Field Video}
	\vspace{-0.7cm}
	\label{fig:algorithm}
\end{figure*}

\noindent
Previous work on temporal alignment of complex dynamic objects has primarily focused on acquisition using multi-view cameras to reliably reconstruct the complete object surface at each frame using shape-from-silhouette and multiple view stereo \cite{BMVCFRanco,StarckH03}.
Robust techniques have been introduced for temporal alignment of the reconstructed non-rigid shape to obtain a 4D model based on tracking the complete surface shape or volume for complex motion \cite{Budd2013,Cagniart2010,StarckH03,Huang2016}. However, these approaches assume a reconstruction of the full non-rigid object surface at each time frame and do not easily extend to 4D alignment of partial surface reconstructions or depth maps.
Recent work obtained reliable temporal alignment of partial surfaces for complex dynamic scenes \cite{Mustafa16ECCV,Tevs12,HaoLi2016ar,Mustafa16}. Dynamic Fusion \cite{Newcombe15} was introduced for 4D modelling from depth image sequences integrating temporal observations of non-rigid shape to resolve fine detail.
However due to the high degree of spatial and temporal regularity and redundancy in the light-field video, existing methods for temporal alignment are not directly applicable.
This paper exploits the spatio-temporal redundancy in light-field video for robust 4D spatio-temporally coherent  representation of dynamic scenes.
Sparse temporal correspondence tracks are obtained across the sequence for each view using temporal feature matching. These sparse temporal correspondence tracks are used to initialize dense scene flow estimation.
A novel sparse-to-dense light-field scene flow is proposed exploiting EPIs to obtain a temporally coherent dense 4D representation of the scene, shown in Figure \ref{fig:algorithm}.
The contributions of the proposed approach include:
\begin{itemize}[topsep=0pt,partopsep=0pt,itemsep=0pt,parsep=0pt] 
	\item Temporally coherent 4D reconstruction of dynamic light-field video;
	\item EPI from sparse light-field video for spatio-temporal correspondence;
	\item Sparse-to-dense light-field scene flow exploiting EPI image information.
\end{itemize}
%
%
\vspace{-0.3cm}
\section{Related work}
\subsection{Light-fields}
With the advent of virtual and augmented reality, light-field capture has been explored for live action immersive virtual reality experiences. 
Bolles et al introduced EPI \cite{Bolles1987} to represent light-field as a volume. This representation has been used effectively for depth estimation and segmentation \cite{Yucer2016,Goldluecke12}.
Methods have been proposed for depth-of-field effects \cite{Levoy2006} and  image-based rendering \cite{Levoy96, Gortler1996}. 
However existing light-field representations have been designed around capturing static objects.
A recent method introduced oriented light-field windows to leverage the EPI information from light-fields and enable more robust and accurate pixel comparisons \cite{Srinivasan_2015_ICCV} to improve the performance for scene flow estimation for a pair of images.
An efficient light-field representation for live action data should account for the high level of redundancy of light-field video and handle multiple resolutions, occlusions, and high variation in depth in complex scenes. 
Another challenging aspect relates to the development of creative tools dedicated to light-field editing \cite{JaraboSIG14}. There is currently a lack of an efficient representation to enable light-field video editing for content production.
In this paper we propose a method to exploit the spatio-temporal redundancy in light-fields. EPI is used to obtain dense scene flow for 4D temporally coherent representation of dynamic light-field video to enable compression and efficient light-field video editing (colour grading, object removal, infilling etc.).
\vspace{-0.2cm}
\subsection{ 4D reconstruction of dynamic scenes}
\vspace{-0.1cm}
For conventional single view depth sequences and multiple view reconstruction of dynamic scenes techniques have been introduced to align sequences using correspondence information between frames. Methods have been proposed to obtain sparse \cite{Mustafa17,Joo_2015_ICCV,Zheng_2015_ICCV,Sundaram2010} and dense \cite{Menze2015CVPR,Zanfir_2015_ICCV,Basha2013} correspondence between consecutive frames for entire sequences.
Existing sparse correspondence methods work independently on a frame-by-frame basis for a single view \cite{Sundaram2010} or multiple views \cite{Joo_2015_ICCV} and require a strong prior initialization \cite{Zheng_2015_ICCV}. Existing feature matching techniques either work in 2D \cite{Sundaram2010} or 3D \cite{Menze2015CVPR} or for sparse \cite{Joo_2015_ICCV,Zheng_2015_ICCV} or dense \cite{Zanfir_2015_ICCV} points. Other methods are limited to RGBD data \cite{Zanfir_2015_ICCV} or stereo pairs \cite{Menze2015CVPR} for dynamic scenes.
Dense matching techniques include scene flow methods. Scene flow techniques \cite{Wedel2011,Basha2013} typically estimate the pairwise surface or volume correspondence between reconstructions at successive frames but do not extend to 4D alignment or correspondence across complete sequences due to drift and failure for rapid and complex motion.
In this paper we propose sparse-to-dense temporal alignment exploiting the high spatio-temporal redundancy in light-fields to robustly align light-field video captured with sparse camera arrays. 
%
%
\vspace{-0.3cm}
\section{Methodology}
\vspace{-0.1cm}
The high spatial and temporal redundancy in light-field video makes it challenging to use in content production for AR/VR. The aim of this work is to obtain a 4D temporally coherent representation of dynamic light-field video exploiting spatio-temporal redundancy. This provides an efficient structured representation for light-field editing and use in immersive VR content production.
\subsection{Overview}
\label{sec:overview}
Given an independent per-frame surface or depth reconstruction from the light-field video captured with a sparse camera array the problem is to simultaneously estimate the temporal correspondence of the input light-field across all views for the entire sequence. 
This is achieved efficiently by estimating the temporal alignment of the reconstructed surface between each time frame and propagating this across all light-field camera views. 
A coarse-to-fine approach is introduced that initially estimates temporal correspondence based on sparse features and then estimates dense scene flow initialised from the sparse features. 
To ensure robust tracking, key-frames are identified as a reference to minimise drift in long-term tracking.
An overview of the 4D temporally coherent reconstruction of light-field video is presented in Figure \ref{fig:algorithm}.

\noindent
\textbf{Light-field video:} 
Light-field video of the dynamic scene is captured using a camera array. Multiple-view stereo is performed to obtain a per frame surface mesh reconstruction \cite{Seitz06}. Throughout this work a sparse 5x4 array with a range of up to 50cm x 50cm is used to capture light-field video.

\begin{figure}[b]
	\centering
	\includegraphics[width=0.99\linewidth]{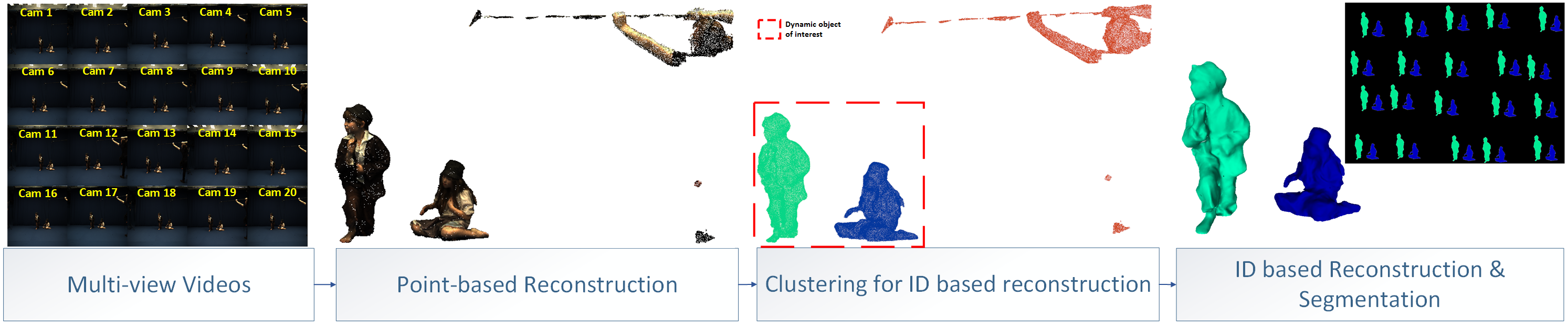}
	\caption{Object identification example on light-field frame}
	\vspace{-0.25cm}
	\label{fig:objectIdentification}
\end{figure}

\noindent
\textbf{Object identification:} The reconstructed point cloud is clustered in 3D \cite{RusuDoctoralDissertation} with each cluster representing a unique foreground object, shown in Figure \ref{fig:objectIdentification}.

\noindent
\textbf{Key-frame detection}: Key-frames are detected for light-field video exploiting redundant spatial information across views to identify a set of unique reference frames for stable long-term tracking. Surface tracking is performed between key-frames to reduce the accumulation of errors in sequential tracking due to large non-rigid motion for long sequences ($\approx \textit{ } 700$ frames).

\noindent
\textbf{Sparse temporal feature correspondence tracks}: Reconstructed 3D points projected at all frames are matched frame-to-frame across the sequence to estimate sparse temporal feature tracks for each dynamic object for each light-field camera view. These sparse temporal correspondence tracks are used to initialize light-field scene flow to handle occlusions and improve robustness.

\noindent
\textbf{Light-field scene flow}: We propose to estimate dense scene flow between images by exploiting EPI information from light-field video based on oriented light-field windows \cite{Srinivasan_2015_ICCV} initialised by the sparse feature correspondence per light-field view at each time instant. This exploits the spatio-temporal redundancy in light-fields to give 2D dense correspondences which are back-projected to the 3D mesh to obtain a 4D spatio-temporally coherent dynamic light-field video representation. 
\subsection{Key-frame detection}
Aligning per-frame reconstruction for long sequences leads to drift due to accumulation of errors in alignment between frames and failure is observed due to large non-rigid motion.
To tackle this problem we detect key-frames across the sequence. Key-frame detection exploits the spatial redundancy in light-field capture by fusing the appearance, distance and shape information across all views ($N_{v}$) in the sparse camera array.
Temporal coherence is introduced between key-frames as explained in Section \ref{sec:sparse} and \ref{sec:dense}.

\noindent
\textbf{Appearance Metric ($M_{i,j}^{c}$):}
This measures appearance similarity between frame $i$ and $j$ for each object region in light-field view $c$. It is the ratio of the number of temporal feature correspondences  $Q_{i,j}^{c}$  to the total number of features in the object region at frame $i$, $R_{i}^{c}$ and $j$, $R_{j}^{c}$: $M_{i,j}^{c} = \frac{2Q_{i,j}^{c}}{R_{i}^{c} + R_{j}^{c}}$

\noindent
\textbf{Distance Metric ($L_{i,j}^{c}$):}
This metric measures the distance between frame $i$ and $j$ for each object in each view $c$, it is defined as: $L_{i,j}^{c} = \frac{j-i}{D_{max}^{c}}$
where $j> i$ and $D_{max}^{c}$ is the maximum number of frames between key-frames for view $c$. This term ensures that the distance between two key-frames does not exceed $D_{max}^{c}$. This is set to $100$.

\noindent
\textbf{Shape Metric ($I_{i,j}^{c}$):} 
Gives the shape overlap between pairs of frames for each object in the light-field video. It is defined as the ratio of the intersection of the aligned segmentation \cite{EvangelidisP08} $h_{i,j}^{c}$ to the union of the area $A_{i,j}^{c}$: $I_{i,j}^{c} = \frac{h_{i,j}^{c}}{A_{i,j}^{c}}$

\noindent
\textbf{Key-frame similarity metric:}
The metrics defined above are used to calculate the similarity between frames as follows: $D_{i,j} = 1 - \frac{1}{3N_{v}}\sum_{c=1}^{N_{v}} ( M_{i,j}^{c} + I_{i,j}^{c} + L_{i,j}^{c})$.
All frames with similarity $>0.75$ are selected as key-frames defined as $K_{i} = \{ {k^{i+1}, k^{i+2}, . . . , k^{N_{f}^{i}}}\}$ where $i=1$ to $N_{k}$ (number of key-frames in light-field video) and $N_{f}^{i}$ is the number of frames between $K_{i}$ and $K_{i+1}$.
%
%
\subsection{Sparse temporal feature correspondence}
\label{sec:sparse}
Numerous approaches have been proposed to temporally align moving objects in 2D using either feature matching or optical flow. However these methods may fail in the case of occlusion, movement parallel to the view direction, large motions and visual ambiguity.
To overcome these limitations we match sparse feature points from all light-field camera views at each time instant for each object. This is used to estimate the similarity between the object surface observed at different frames of the light-field video for key-frame detection and subsequently to initialize dense light-field scene flow between frames.
\begin{figure}[t]
	\centering
	\includegraphics[width=0.99\linewidth]{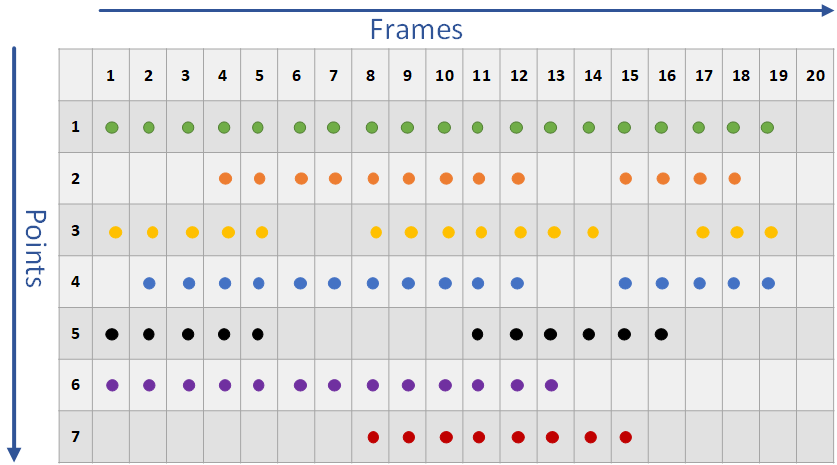}
	\caption{Example of sparse feature correspondence tracks}
	\vspace{-0.1cm}
	\label{fig:sparseFeatureTracks}
\end{figure}
%

\noindent
3D points corresponding to each dynamic object are projected at each frame to ensure spatial light-field coherence. The features for each light-field camera view are defined as:
$F^{c}_{i} = \{ {f^{c}_{1}, f^{c}_{2}, . . . , f^{c}_{R_{i}^{c}}}\}$, where $c={1 \text{ to } N_{v}}$ and $N_{v} = 20$ in our case. $R_{i}^{c}$ are the 3D points visible at each frame $i$.
Nearest neighbour matching is used to establish matches between features. The ratio of the first to second nearest neighbour descriptor matching score is used to eliminate ambiguous matches ($ratio < 0.85$).
This is followed by a symmetry test which employs the principle of forward and backward match consistency to remove erroneous inconsistent correspondences.
To further refine the sparse matching and eliminate outliers we enforce local spatial coherence in the matching.
For matches in an $m \times m$ ($m = 11$) neighbourhood of each feature we find the average Euclidean distance and constrain the match to be within a threshold 
($\pm \eta < 2 \times \text{Average Euclidean distance}$). 

\noindent
Sparse temporal correspondence tracks are obtained by performing exhaustive matching between all frames $N_{f}^{i}$ for each key-frame for the entire sequence.
Feature matching is performed between frames such that features at view $c$ frame $i$, $F^{c}_{i}$ are matched to features at view $c$ to frames $j = \{ {i+1, . . . , N_{f}^{i}}\}$. This gives us correspondences for all the frames $N_{f}^{i}$ with key-frame $K_{i}$. Point tracks are constructed from this correspondence information for key-frame $K_{i}$.
The same process is repeated for key-points which are not part of point tracks at the corresponding key-frame for all frames $j = \{ {i+1, . . . , N_{f}^{i}}\}$. Any new point-tracks are added to the list of point tracks for key-frame $K_{i}$.
The exhaustive matching between frames per key-frame handles reappearance and disappearance of points due to occlusion or object movement. An example of sparse feature correspondence tracks in shown in Figure \ref{fig:sparseFeatureTracks}. Frame $1$ is a key-frame and each point track is represented by a unique colour with missing points showing partial feature correspondence tracks.
Sparse correspondences are also obtained between all key-frames to initialise dense light-field flow.
\begin{figure*}
	\centering
	\includegraphics[width=0.99\linewidth]{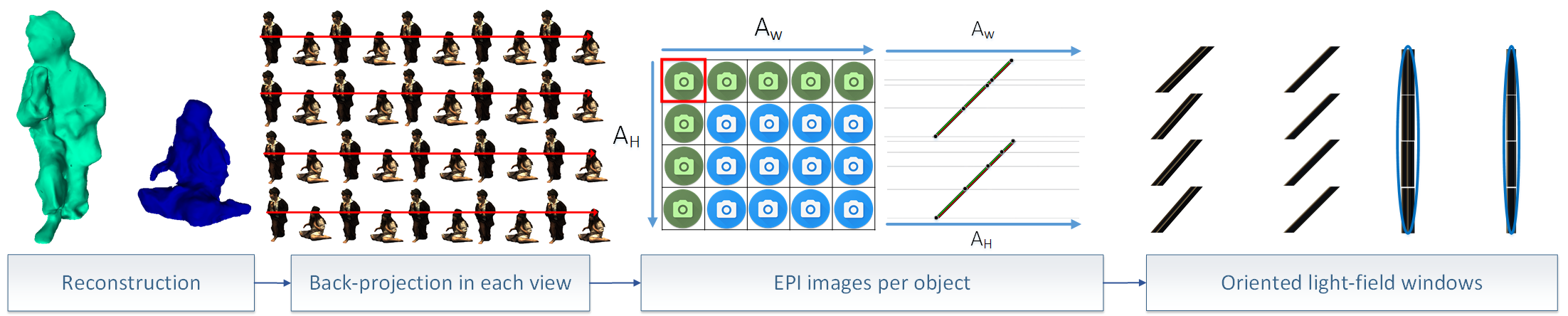}
	\caption{An example of EPI representation of sparse light-field (blue ellipse represents Gaussian weighted window)}
	\vspace{-0.5cm}
	\label{fig:epi}
\end{figure*}
%
\vspace{-0.15cm}
\subsection{Light-field scene flow}
\label{sec:dense}
Dense temporal correspondence is estimated using a novel light-field scene flow method using pairwise dense correspondence which is estimated using the light-field oriented window matching. This combines information across all light-field views to achieve robust temporal correspondence.
The sparse feature correspondences provides a robust initialisation of the proposed dense flow for large non-rigid shape deformation in the light-field video.

\noindent
The high-level of inherent redundancy in the light-field images is used to improve the robustness and reliability of dense scene flow by using oriented light-field windows on the EPI for matching. This has been shown to improve the temporal correspondences for scene flow \cite{Srinivasan_2015_ICCV}.
In this paper we propose a novel method to obtain EPIs from sparse light-field data to improve the quality of dense scene flow. Our approach uses oriented light-field windows \cite{Srinivasan_2015_ICCV} to represent surface appearance information for matching as illustrated in Figure \ref{fig:epi}.
\vspace{-0.35cm}
\subsubsection{Epipolar Plane Image (EPI)}
\begin{figure}[t]
	\centering
	\includegraphics[width=0.99\linewidth]{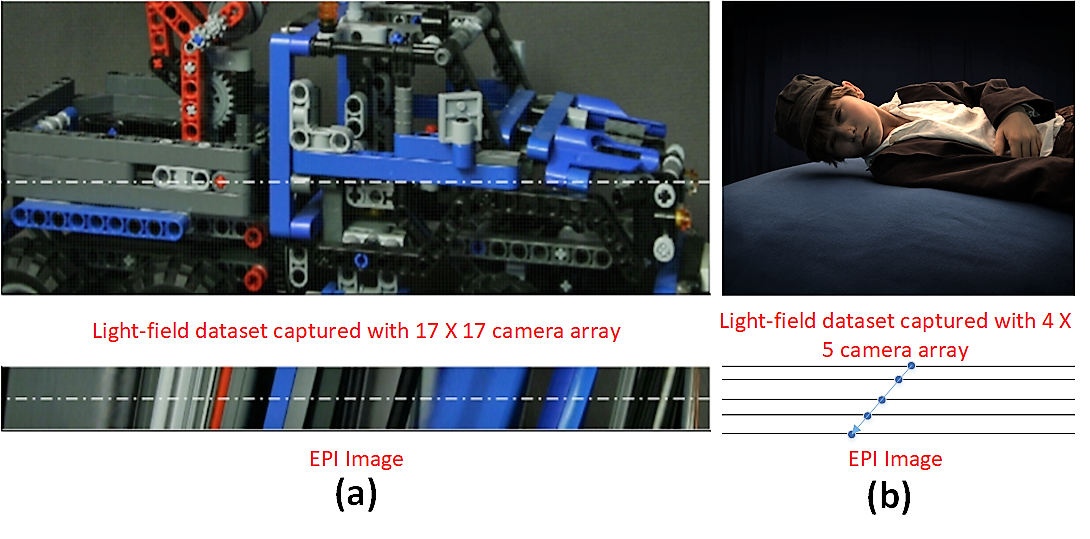}
	\vspace{-0.2cm}
	\caption{Illustration of EPI from (a) dense Stanford dataset \cite{StanfordGraphicsLaboratory2008} (EPI from \cite{Goldluecke12}) and (b) sparse light-field capture}
	\vspace{-0.1cm}
	\label{fig:epidense}
\end{figure}

Traditional light-field data is captured with a plenoptic camera or a dense camera array typically capturing 200-300 views \cite{StanfordGraphicsLaboratory2008}. An example dataset is illustrated in Figure \ref{fig:epidense} (a) captured with 289 images. Given a dense set of views the EPI provides a representation for estimating correspondence across views from the regular structure of the image i.e. slant lines correspond to the same surface point. The 2D slice of such representation has the same width as the captured image and height is given by the number of views in a camera array row ($17$ in Figure \ref{fig:epidense} (a)). With dense sampling the disparity between adjacent views is typically sub-pixel giving slant lines in the EPI. Approaches have been proposed to estimate depth and segmentation from this dense EPI representation \cite{Yucer2016,Goldluecke12}.
However for sparse camera sampling the disparity between views may be several pixels making it difficult to directly establish surface correspondence from the EPI, Figure \ref{fig:epidense} (b). In our case of sparse views with just $20$ cameras, the height of this EPI reduces to $5$ pixels which makes it challenging to utilize the regularity of the information from EPI obtained from light-fields.

\noindent
In this paper we aim to use the EPIs to introduce spatio-temporal coherence in dynamic light-field video. We propose a novel method to create an EPI parametrized representation from sparse light-field capture. We assume that the calibration is known. All the images at each time instant are undistorted and rectified with respect to the reference camera. The depth information at each time instant is used to resample the light-field to create an EPI representation of sparse light-field data. The algorithm to obtain EPI from image, calibration and depth information is illustrated in Figure \ref{fig:epi}; the stages are as follows:
\begin{itemize}[topsep=0pt,partopsep=0pt,itemsep=0pt,parsep=0pt] 	
	\item The dense point-cloud $P$ of each dynamic object is projected on the undistorted and rectified images. $B^{c}_{i,j}$ is the set of points in view $c$.  For each row $j$ of the projected points on the dynamic object a 2D EPI is obtained.
	\item The size of 2D EPI is set to $W \times H$, where $W$ is the width of input images and $H = N_{w} \times \mu$ is estimated corresponding to the number of cameras in each row ($N_{w}$). $\mu$ is a constant introduced to increase the distance between views due to sparse camera sampling. It is set to $50$ in our case to maximize performance.
	For each row of the camera array, $H_{o}$ 2D EPIs are obtained for each dynamic object, where $H_{o}$ is the height of the object.
	The corresponding projected points $B^{c}_{i,j}$ for each point $P_{i,j}$ are plotted on the 2D EPI with $x$ coordinates the same as that of input images and $y$ coordinates are estimated using the translation information from the calibration, defined as: $	\frac{H \text{ } X \text{ Distance between consecutive cameras}}{\text{Maximum distance between pair of cameras}}$
	\item  Given the set of image samples corresponding to a given surface point we fit a line in the EPI due to imperfect camera array. A scene point is represented by a line in the 2D EPI. A similar process is repeated for the entire point-cloud $P$ to obtain multiple 2D EPIs. Given a 2 dimensional camera array there are 2 sets of EPIs obtained by resampling along epipolar lines in the vertical and horizontal directions
\end{itemize}
The EPIs obtained from sparse camera arrays are used to constrain dense light-field flow using oriented light-field windows.
\vspace{-0.3cm}
\subsubsection{Oriented light-field windows}
\vspace{-0.1cm}
Oriented light-field windows exploit the regularity information in the EPI to enable more robust and accurate pixel comparisons \cite{Srinivasan_2015_ICCV} over spatial windows which suffer from defocus blur and loss of precision.
In this paper we propose to use these oriented light-field windows on the EPIs obtained from the sparse light-field data to estimate the light-field scene flow temporal correspondence.

\noindent
Each scene point can be represented by an oriented window in the light-field ray space. The shear of the ray is related to the depth of the 3D point and the size of the window is defined by spatial and angular Gaussian weights. However the ray moves with object motion, hence there is a need to account for shear and translation. The oriented light-field window corresponding to a scene point for a light-field, $L$ is computed as follows:
\vspace{-0.25cm}
\begin{equation} \label{eq:lfcost}
O_{d,x0,y0} (x, y, u, v) = \left ( W * S_{d} * T_{x0,y0} \right )\left [  L \right ]
\vspace{-0.25cm}
\end{equation}
where, $x_{0},y_{0}$ is the centre of the window, $x,y$ represents the image plane and $u,v$ represents the camera plane. $S_{d}$ is the shear operation for depth $d$, defined as $S_{d}\left [ L \right ] = L_{d}(x,y,u,v) = L(x + u \left ( 1 - \frac{1}{d} \right ), y + v \left ( 1 - \frac{1}{d} \right ),u,v) $, $T_{x0,y0}$ is the translation operator defined as, $T_{x0,y0} \left [ L \right ] = L_{x0,y0}(x, y, u, v) = L(x + x0, y + y0, u, v)$ and $W$ is the Gaussian weighted windows defined as: $W \left [ L \right ] = L(x, y, u, v) N (x, y; \sigma^{2}_{xy})A(u,v)$. $N()$ is the 2D Gaussian distribution windows and $A(u,v)$ is the corresponding row and column of the camera array (shown as green cameras in Figure \ref{fig:epi}, $3^{rd}$ step)  of the reference light-field view (($u_{c}, v_{c} $) shown in red box), defined as:
\vspace{-0.3cm}
\begin{equation*}
A(u,v) = \left\{\begin{matrix}
1, & \text{ if } v = v_{c} \text{ or } u = u_{c} \\ 
0, & otherwise
\end{matrix}\right.
	\vspace{-0.2cm}
\end{equation*}
This formulation could be easily extended to dense camera array.  
We propose to use this oriented light-field window to robustly estimate the dense scene flow for light-field video.
\begin{figure*}
	\centering
	\includegraphics[width=0.99\linewidth]{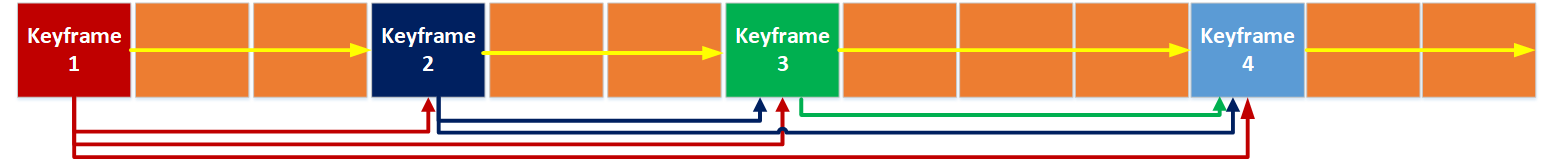}
	\caption{Illustration of dense correspondence between frames and key-frames for full sequence 4D alignment: arrows indicate sparse correspondence tracks and dense correspondence}
	\vspace{-0.5cm}
	\label{fig:keyframe}
\end{figure*}
\vspace{-0.7cm}
\subsubsection{Dense light-field scene flow}
\begin{figure}
	\centering
	\includegraphics[width=0.99\linewidth]{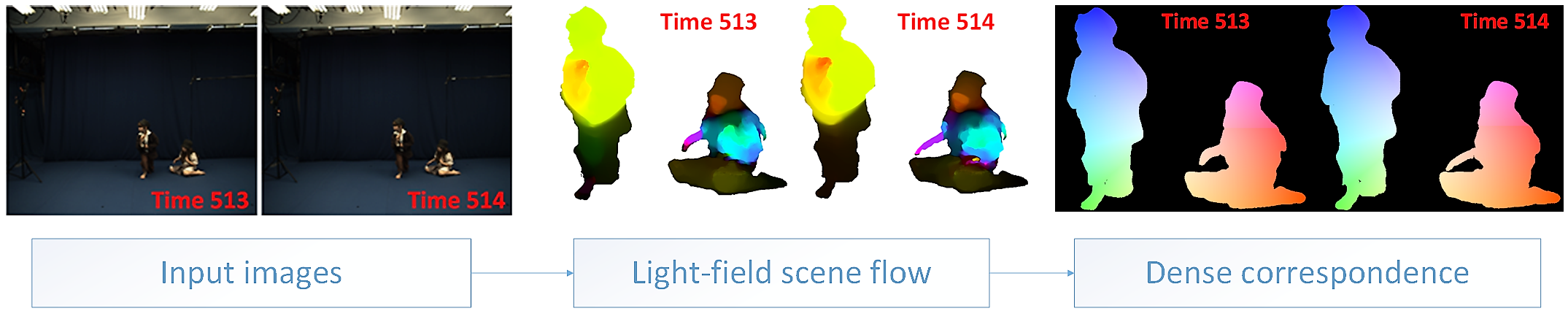}
	\caption{Light-field scene flow for two consecutive frames}
	\label{fig:flow}
\end{figure}

Oriented light-field windows are used to estimate flow between consecutive frames and between key-frames, Figure \ref{fig:keyframe}, using the sparse temporal feature correspondence as initialization for each view. Flow is estimated on the object region for each light-field view, as illustrated in Figure \ref{fig:flow}, to obtain dense temporal correspondence.

\noindent
The flow is formulated as a translation of each pixel location ${p} = (x_p, y_p)$ in image $I$ by $m_p = (\delta x_p, \delta y_p)$ in time. We formulate the computation of flow $\textit{\textbf{M}}$ per view for each dynamic object by minimization of the cost function:
\vspace{-0.25cm}
\begin{equation} \label{eq:costfunction}
E(\textit{\textbf{M}}) = \sum_{{p} \in I} \lambda _{L}E_{L}(p,m_p) + \lambda _{C}E_{C}(p,m_p)  +  \lambda _{R}E_{R}(p, m_p)
\vspace{-0.15cm}
\end{equation}
The cost consists of three terms: the light-field consistency $E_{L}$ for the oriented light-field window alignment; the appearance term $E_{C}$ for brightness coherency; and the regularization term $E_{R}$ to avoid sudden peaks in flow and maintain the consistency. Colour and regularization terms are common to optical flow problems \cite{Tao2012} and the light-field consistency is introduced for the first time to improve dense flow for sparse light-field video. \\
\textbf{Light-field Consistency:} The 2D EPIs obtained from the sparse light-field views are used to define oriented light-field windows for each scene point. These windows encapsulate the observed multi-view light-field appearance of the corresponding surface point and can be matched over time to estimate the temporal correspondence, defined as: 
\vspace{-0.25cm}
\begin{equation*} \label{eq:lf} 
\begin{aligned} 
E_{L}(p,m_p) = \left \| O_{d_{t},p} (p, u, v, t) - O_{d_{t+1}, p + m_p} (p, u, v, t+1) \right \| ^{2} \text{ }
\end{aligned}
\vspace{-0.25cm}
\end{equation*}
where $O_{d_{t},m}()$ is the oriented light-field window at time $t$ with depth $d_{t}$ as defined in Equation \ref{eq:lfcost}.
\\
\textbf{Appearance Consistency:} This adds the brightness consistency assumption to the cost function generalized for all $N_{v}$ light-field cameras for both time steps. This term is obtained by integrating the sum of three penalizers over the reference image domain. $e_{C}^{T}()$ penalizes deviation from the brightness constancy assumption in time for same views; $e_{C}^{V}()$ penalizes deviation from the brightness constancy assumption between the reference view and each of the other views at time $t + 1$. $e_{C}^{S}()$ forces the flow to be close to nearby sparse temporal correspondences.
\vspace{-0.25cm}
\begin{eqnarray*}
E_{C}({p,m_p}) = e_{C}^{T}({p,m_p}) + e_{C}^{V}({p,m_p}) + e_{C}^{S}({p,m_p}) \\
e_{C}^{T}({p,m_p}) =\sum_{i = 1}^{N_{v}} \left \| (I_{i}(p,t) - I_{i}(p+ m_p, t+1)) \right \| ^{2} \\
e_{C}^{V}({p,m_p}) =\sum_{i = 2}^{N_{v}} \left \| (I_{1}(p,t) - I_{i}(p+ m_p, t+1)) \right \| ^{2} \\
e_{C}^{S}({p,m_p})   =\sum_{i = 1}^{N_{v}} e_{C}^{S} = \left\{\begin{matrix} 0 & \text{if } p \in N \\ \infty & otherwise \end{matrix}\right.
\vspace{-0.5cm}
\end{eqnarray*}
where $I_{i}(p,t)$ is the intensity at point $p$ and time $t$ in camera $i$.
This term denotes that the flow vector $m$ is located within a window from a sparse constraint at $p$ and it forces the flow to approximate the sparse 2D temporal correspondence tracks.
\\
\textbf{Regularization:} This penalizes the absolute difference of the flow field to enforce motion smoothness and handle occlusions and areas with low confidence.
\vspace{-0.2cm}
\begin{eqnarray*}
E_{R}({p,m_p}) = \sum_{p,q \in N_{p}} \left \| \Delta m \right \|^{2} (\lambda_{R}^{L} e_{R}^{L} (p, q, m_{p}, m_{q}) + \\
\lambda_{R}^{C} e_{R}^{C} (p, q, m_{p}, m_{q}) ) \\
e_{R}^{L} (p, q, m_{p}, m_{q}) = \underset{q \in N_{p}}{mean} E_{L}({q,m_q}) - \underset{q \in N_{p}}{min} E_{L}({q,m_q})
\\
e_{R}^{C} (p, q, m_{p}, m_{q}) = \underset{q \in N_{p}}{mean} E_{C}({q,m_q}) - \underset{q \in N_{p}}{min} E_{C}({q,m_q})
\vspace{-1cm}
\end{eqnarray*}
where $\Delta m = m_{p} - m_{q}$ and we compute $e_{R}^{L}$ and $e_{R}^{C}$ as the minimum subtracted from the mean data energy within the search window $N_{p}$ for each pixel $p$.
\\
\textbf{Occlusions:}
To detect occlusions, we compare the forward flow from $t$ to $t + 1$, and the backward flow from $t + 1$ to $t$. Occluded pixels are robustly indicated by large differences in the forward/backward motion and excluded as outliers.
\\
\textbf{Optimization:}
For each pixel $p$, the energy is estimated as defined in eq. \ref{eq:costfunction} on a window $N_{p}$, as in the SimpleFlow algorithm \cite{Tao2012} to estimate the flow vector $m$. The optical flow is optimized over a multi-scale pyramid with warping between pyramid levels, resulting in a coarse-to-fine strategy that allows the estimation of large displacements by minimizing the equation defined as: $F(\textit{\textbf{M}}) = \argmin E(\textit{\textbf{M}})$
\vspace{-0.25cm}
\subsubsection{4D temporally coherent light-field video}
The estimated dense flow for each view is back projected to the 3D visible surface to establish dense 4D correspondence between frames ($N_{f}^{i}$) and between key-frames $K_{i}$ as seen in Figure \ref{fig:keyframe} to obtain 4D temporally coherent light-field video.
Dense 4D correspondence is first obtained for the light-field view with maximum visibility of 3D points. To increase surface coverage correspondences are added in order of visibility of 3D points for different sparse light-field views.
Dense temporal correspondence is propagated to new surface regions as they appear using the sparse feature correspondence tracks and respective dense light-field scene flow.
Dense scene flow is estimated between key-frames for robust long-term surface tracking.
%
%
\begin{table*}
	\centering
	\begin{tabular}{l|c|c|l|l|c|c}
	\textbf{Datasets} & \multicolumn{1}{l|}{\textbf{Sequence length}} & \multicolumn{1}{l|}{\textbf{No. of views}} & \textbf{Shot level} & \textbf{Resolution} & \multicolumn{1}{l|}{\textbf{Key-frames}} & \multicolumn{1}{l}{\textbf{Avg. sparse tracks}} \\ \hline
	Walking & 667 & 20 & Far & 2448 X 2048 & 15 & 1934 \\
	Sitting & 694 & 20 & Mid-level & 2448 X 2048 & 13 & 1046 \\
	Wakingup & 270 & 20 & Close-up & 2448 X 2048 & 7 & 2083 \\
	Running & 140 & 20 & Far & 2448 X 2048 & 5 & 1278 \\
	Magician & 353 & 20 & Close-up & 2448 X 2048 & 6 & 1312
	\end{tabular}
	\caption{Properties of all datasets: Key-frames gives the number of key-frames detected for the dynamic sequence and Avg. sparse tracks gives the number of sparse temporal correspondence tracks averaged over the entire sequence for each object}
	\label{datasets}
	\vspace {-0.3cm}
\end{table*}
\begin{figure*}[t]
	\centering
	\includegraphics[width=0.99\linewidth]{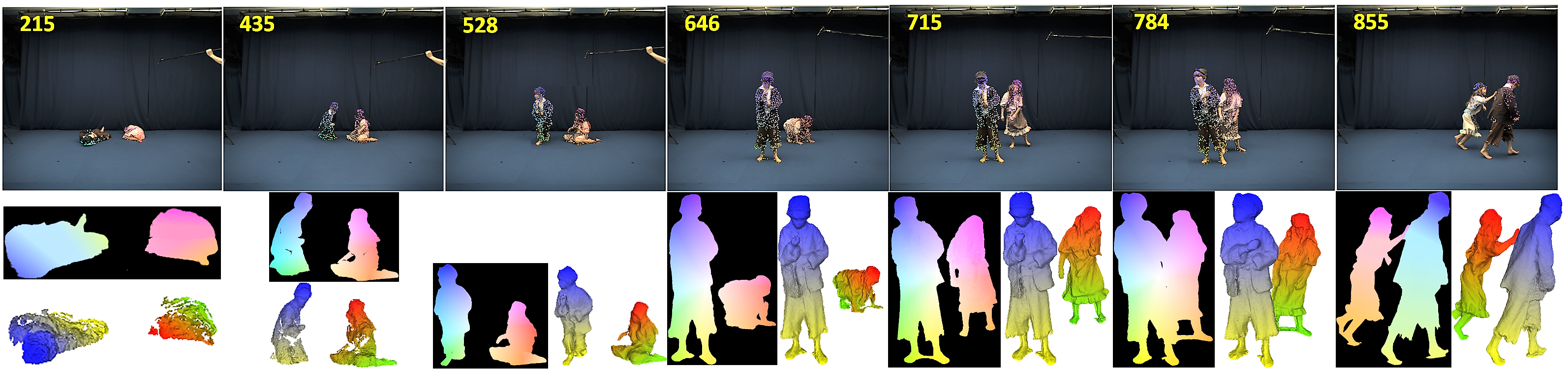}
	\caption{Dense 2D and 4D correspondence for Walking sequence across key-frames}
	\vspace{-0.5cm}
	\label{fig:full}
\end{figure*}
\begin{figure}
	\centering
	\includegraphics[width=0.99\linewidth]{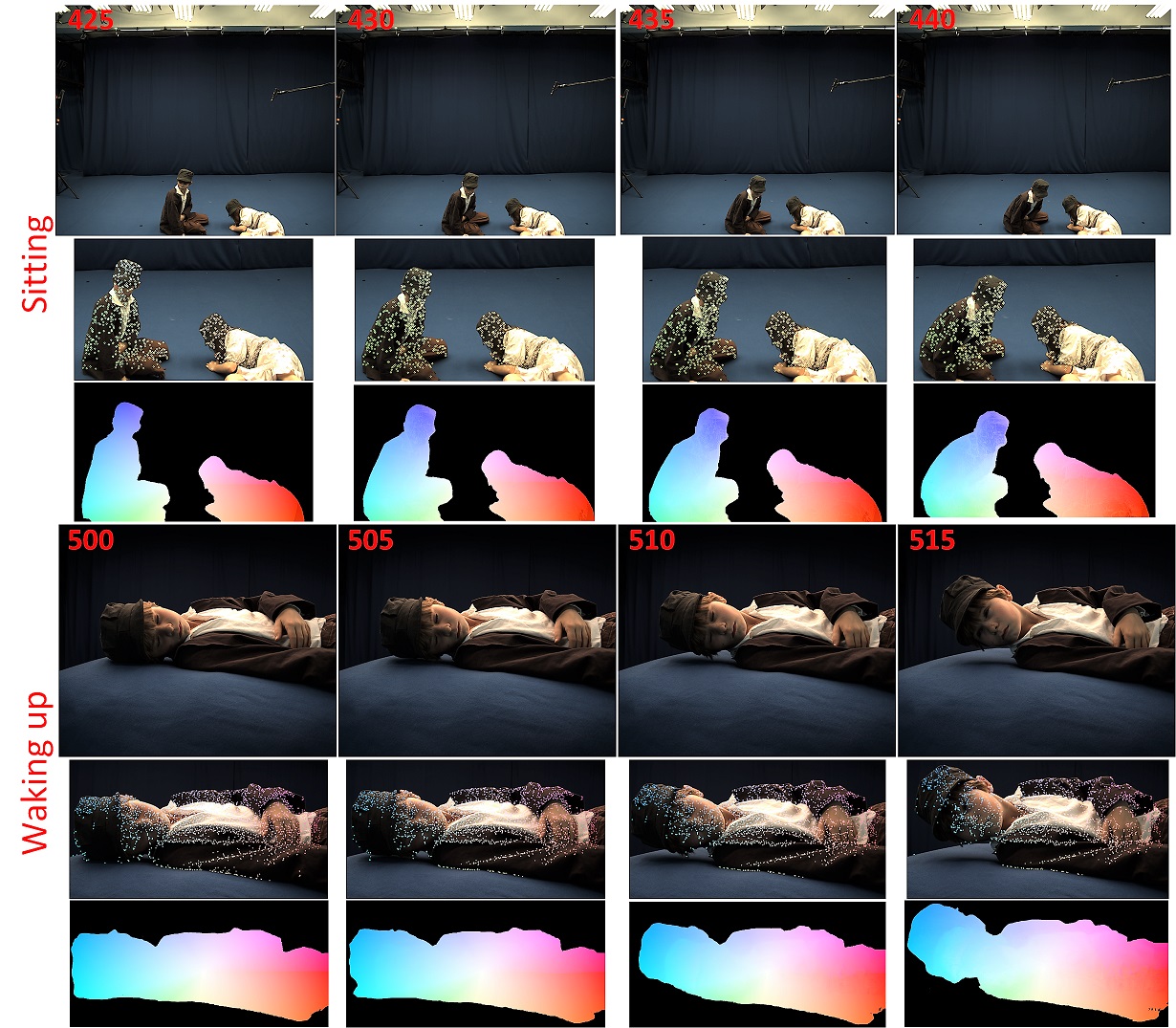}
	\caption{Sparse temporal correspondences and dense flow results on 2 light-field sequences: Sitting and Waking up}
	\label{fig:results}
\end{figure}
\begin{figure}
	\centering
	\includegraphics[width=0.99\linewidth]{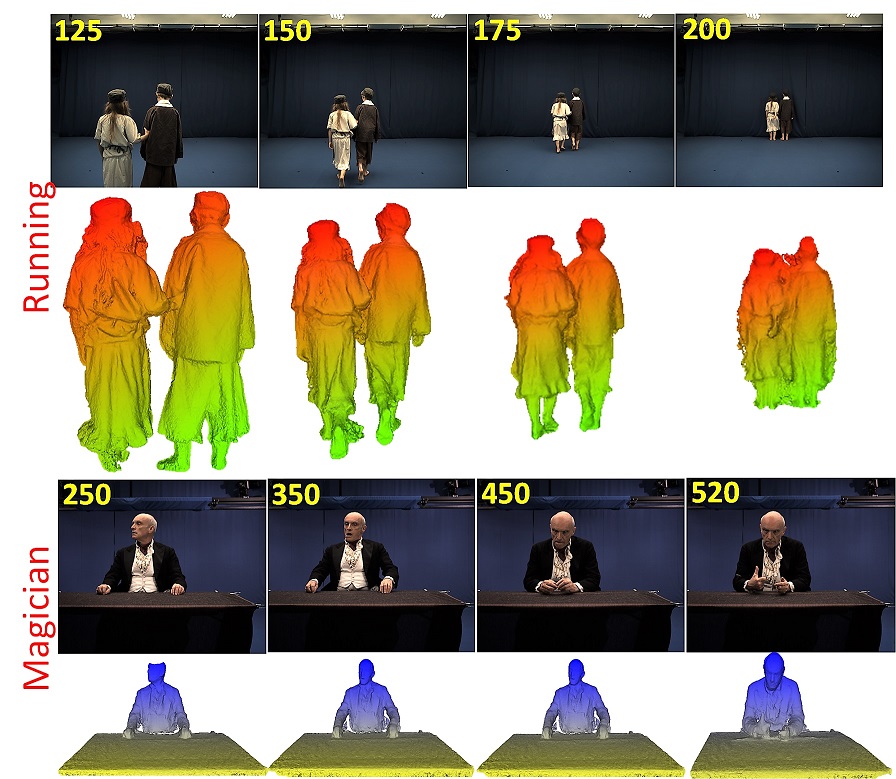}
	\caption{4D temporal alignment between frames for Walking and Magician dataset}
	\label{fig:3Dflow}
\end{figure}
\begin{figure}
	\centering
	\includegraphics[width=0.99\linewidth]{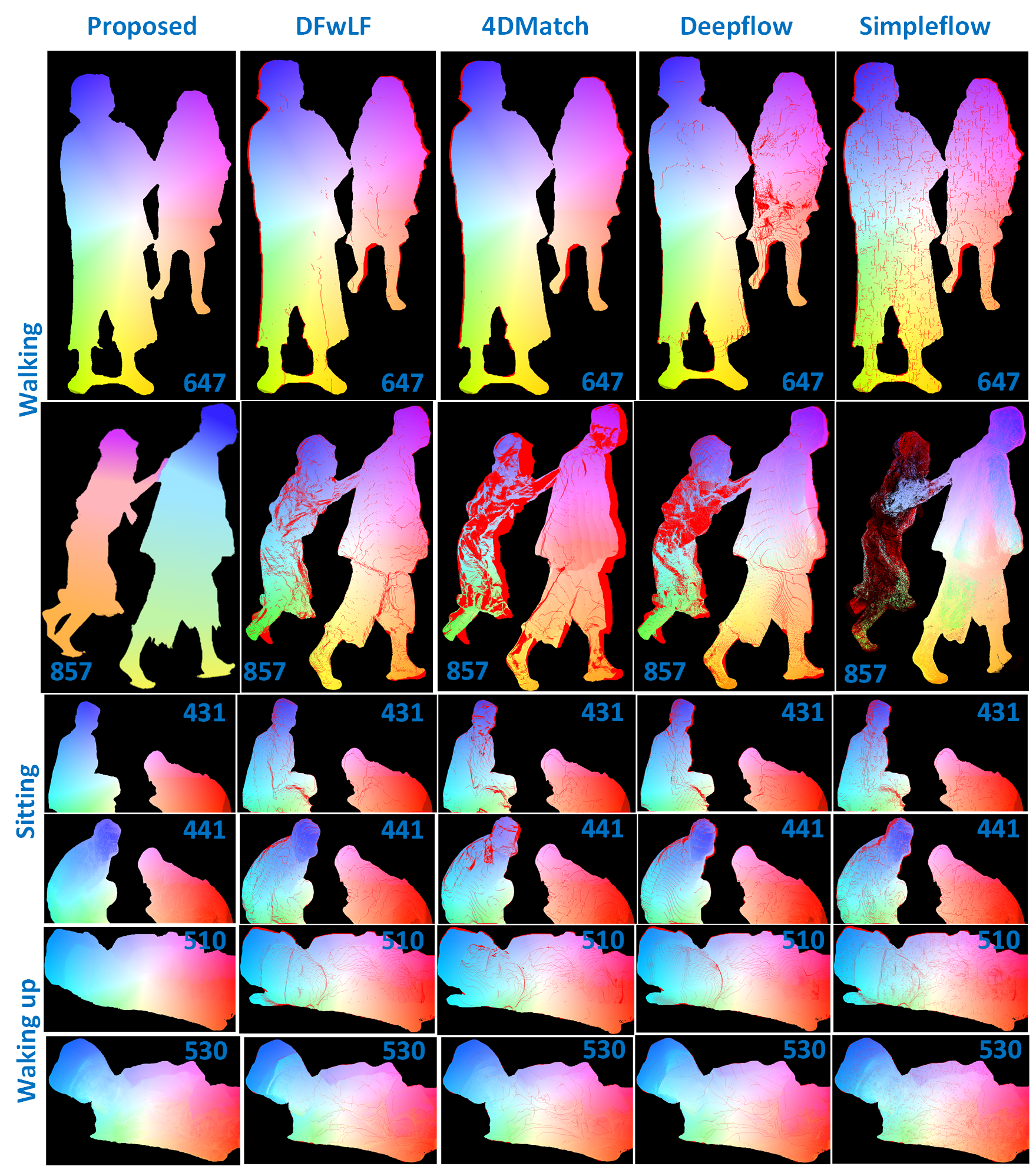}
	\caption{Dense flow comparison results on different light-field sequences}
	\label{fig:denseflow}
\end{figure}
\begin{figure}
	\centering
	\includegraphics[width=0.99\linewidth]{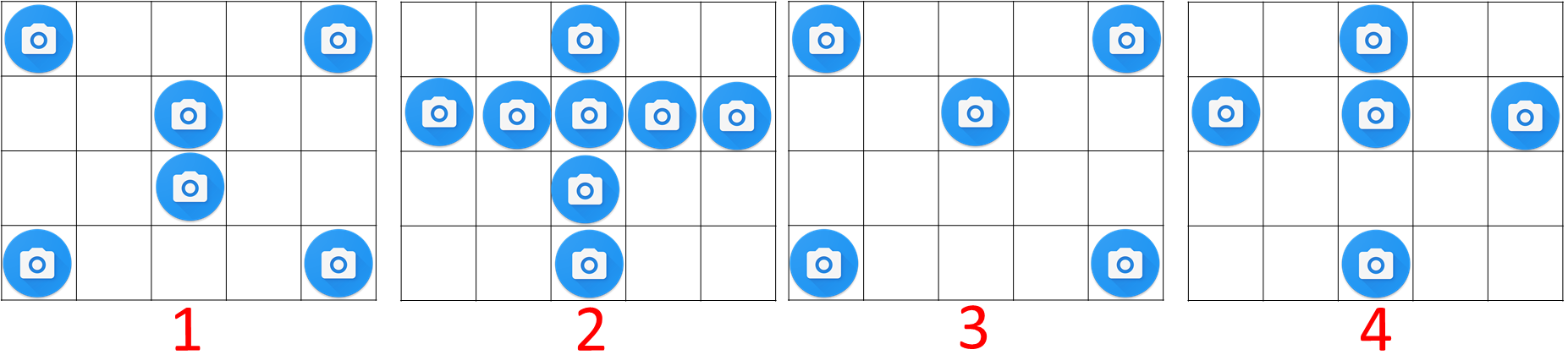}
	\caption{Different light-field camera array configurations}
	\label{fig:config}
\end{figure}
%
%
%
\section{Results and Performance Evaluation}
\label{sec:results}
The proposed approach is tested on various light-field captures. The properties of the evaluation datasets are presented in Table \ref{datasets}. Algorithm parameters set empirically are constant for all results.

\noindent
Sparse and dense correspondence are obtained on the sparse light-field dynamic data and the colour coded results are shown in Figure \ref{fig:full} for Walking dataset and in Figure \ref{fig:results} for Sitting and Waking up dataset using the method explained in Section \ref{sec:overview}.
To illustrate the 2D dense alignment the silhouette of the dense mesh on key-frames is colour coded and the colours are propagated between frames using dense scene flow explained in Section \ref{sec:dense}.
Results of the proposed 4D temporal alignment, illustrated in Figure \ref{fig:3Dflow} shows that the colour of the points remains consistent between frames. 
The proposed approach is qualitatively shown to propagate the correspondences reliably over the entire light-field video for complex dynamic scenes with large non-rigid motion.\\
\begin{table}
	\centering
	\begin{tabular}{l|l|l|l|l|l}
		Datasets & Prop. & DFwLF & 4DM & DF & SF \\ \hline
		Walking & \textbf{0.45} & 0.59 & 0.58 & 0.81 & 1.05 \\ 
		Sitting & \textbf{0.51} & 0.73 & 0.71 & 1.13 & 1.83 \\ 
		Waking up & \textbf{0.39} & 0.56 & 0.53 & 0.89 & 1.17 \\ 
		Running & \textbf{0.65} & 0.87 & 0.92 & 1.23 & 1.95 \\ 
		Magician & \textbf{0.59} & 0.82 & 0.83 & 1.05 & 1.67 \\ 
	\end{tabular}
	\caption{Silhouette overlap error for all the datasets. Prop. represents proposed approach, 4DM is 4DMatch, DF is Deepflow and SF is Simpleflow}
	\label{soe}
\end{table}
\begin{table}[]
	\centering
	\begin{tabular}{l|c|c|l|l}
		& \multicolumn{2}{c|}{\textbf{Frame-to-frame}} & \multicolumn{2}{c}{\textbf{Keyframe-to-frame}} \\ \hline
		\multicolumn{1}{l|}{\textbf{Methods}} & \multicolumn{1}{l|}{\textbf{Mean}} & \multicolumn{1}{l|}{\textbf{S.D.}} & \textbf{Mean} & \textbf{S.D.} \\ \hline
		\multicolumn{1}{l|}{Proposed} & \textbf{ 3.78} & 1.45 & \textbf{4.34} & 1.74 \\ 
		\multicolumn{1}{l|}{DFwLF} &  5.93 & 2.60 & 7.41 & 3.73  \\ 
		\multicolumn{1}{l|}{4DM} &  5.30 & 2.12 & 6.95 & 3.12 \\ 
		\multicolumn{1}{l|}{DF} &  6.28 & 3.77 & 15.79 & 6.36 \\ 
		\multicolumn{1}{l|}{SF}&  7.82 & 4.31 & 21.92 & 8.45 \\ 
	\end{tabular}
	\caption{Temporal coherence evaluation for Walking dataset against existing methods: S.D. is the standard deviation}
	\label{temporal}
\end{table}
\textbf{Qualitative evaluation:}
For comparative evaluation we use:(a) state-of-the-art dense flow algorithm Deepflow \cite{deepflow}; (b) dense flow without light-field consistency (DFwLF) in eq. \ref{eq:costfunction}; (c) a recent algorithm for alignment of partial surfaces (4DMatch) \cite{Mustafa16ECCV} and (d) Simple flow \cite{Tao2012}.
Qualitative results against DFwLF, 4DMatch, Deepflow and Simpleflow shown in Figure \ref{fig:denseflow} indicate that the propagated colour map does not remain consistent across the sequence for large motion as compared to the proposed method (red regions indicate correspondence failure).\\
\textbf{Quantitative evaluation:}
For quantitative evaluation we compare the silhouette overlap error (SOE). Dense correspondence over time is used to create propagated mask for each image. The propagated mask is overlapped with the silhouette of the projected surface reconstruction at each frame to evaluate the accuracy of the dense propagation. The error is defined as: $SOE = \frac{1}{M  N}\sum_{i = 1}^{N}\sum_{c = 1}^{M} \frac{\text{Area of intersection}}{\text{Area of back-projected mask}}$. Evaluation against the different techniques is shown in Table \ref{soe} for all datasets. 
As observed the silhouette overlap error is lowest for the proposed approach showing relatively high accuracy.

\noindent
We evaluate the temporal coherence across Walking sequence, by evaluating the variation in appearance for each scene point between frames and between key-frames and frames for state-of-the-art methods, defined as: $\sqrt{\frac{\Delta r^{2} + \Delta g^{2} + \Delta b^{2}}{3}}$, where $\Delta$ is the difference operator. Evaluation shown in Table \ref{temporal} against state-of-the-art methods demonstrates the stability of long term temporal tracking for proposed method.
Evaluation of the proposed method against dense flow without light-field consistency (DFwLF) demonstrates the usefulness of information from the EPIs in the dense flow in Section \ref{sec:dense}.\\
\textbf{Light-field camera array configuration:}
We evaluate the performance of the proposed 4D temporal alignment with different light-field camera configurations shown in Figure \ref{fig:config}. 
The completeness of the 3D points at each time instant for all camera configurations as observed in Table \ref{completeness} is defined as: $ \frac{100}{M  N}\sum_{i = 1}^{N}\sum_{c = 1}^{M} \frac{\text{Number of 3D points propagated in each configuration}}{\text{Number of 3D points propagated with full camera array}}$. The evaluation demonstrates a drop in 3D correspondence with the reduction in number of cameras in different camera configurations, specially for close-up shots (Wakeup and Magician). However configuration $1$ and $3$ with cameras in the corners provide a better coverage compared to $2$ and $4$.\\
\begin{table}
	\centering
	\begin{tabular}{c|l|l|l|l|l}
		\textbf{Config.} & Walk & Sit & Wake & Run & Magician \\ \hline
		\textbf{1} & 95.15 & 92.58 & 89.64 & 91.03 & 90.82 \\ 
		\textbf{2} & 91.33 & 86.73 & 86.98 & 88.78 & 89.90 \\ 
		\textbf{3} & 90.76 & 85.15 & 86.05 & 87.21 & 86.21 \\
		\textbf{4} & 87.82 & 82.40 & 85.16 & 86.56 & 80.91 \\ 
	\end{tabular}
	\caption{Completeness of dense 3D correspondence averaged over the entire sequence in \% for different camera configurations}
	\label{completeness}
\end{table}
\noindent
\textbf{Limitations:} The proposed method fails for fast spinning objects; scenes with uniform appearance and highly crowded dynamic environments. This is due to the failure of sparse and dense correspondence due to high amibiguity. 
%
\vspace{-0.55cm}
\section{Conclusion}

This paper introduced the first algorithm to obtain a 4D temporally coherent representation of dynamic light-field video.
A novel method to obtain EPIs from sparse light-field video for spatio-temporal correspondence is proposed.
Sparse-to-dense light-field scene flow is introduced exploiting information from the EPIs.
Dense correspondence is fused spatially for 4D temporally coherent light-field video.
The proposed approach is evaluated on various light-field sequences of complex dynamic scenes with large non-rigid deformations to obtain a temporally consistent 4D representation and demonstrating accuracy of the resulting 4D alignment.
4D Light-field video provides a spatio-temporally coherent representation to support subsequent light-field video compression or editing to replicate the functionality of conventional video editing allowing the propagation of edits both spatially across views and temporally across frames.

\noindent \textbf{Acknowledgments:} This research was supported by the InnovateUK grant for Live Action Lightfields for Immersive Virtual Reality Experiences (ALIVE) project (grant 102686).

{\small
\bibliographystyle{ieee}
\bibliography{egbib}
}

\end{document}